\relax
\documentclass[letterpaper]{article}
\usepackage[final]{nips_2016}
\usepackage{subfigure}
\usepackage{amsmath,amssymb,amsthm}
\usepackage{graphicx}
\usepackage{natbib}
\usepackage{color}
\usepackage{algorithm,algorithmic} 
\usepackage{nicefrac}
\usepackage{caption}
\usepackage{float}

\usepackage{adjustbox}
\usepackage{booktabs}
\usepackage{bm}

\DeclareMathOperator*{\argmin}{\arg\!\min}
\DeclareMathOperator*{\argmax}{\arg\!\max}

\newlength\myindent
\title{Model-based Adversarial Imitation Learning}
\author{Nir Baram, Oron Anschel, Shie Mannor \\
Electrical Engineering Department,\\
Technion - Israel Institute of Technology,\\
Haifa 32000, Israel\\
\texttt{\{nirb@campus, oronanschel@campus, shie@ee\}.technion.ac.il}
}	
\begin{document} 

\maketitle
\begin{abstract}
Generative adversarial learning is a popular new approach to training generative models which has been proven successful for other related problems as well. The general idea is to maintain an oracle $D$ that discriminates between the expert's data distribution and that of the generative model $G$. The generative model is trained to capture the expert's distribution by maximizing the probability of $D$ misclassifying the data it generates. Overall, the system is \emph{differentiable} end-to-end and is trained using basic backpropagation.
This type of learning was successfully applied to the problem of policy imitation in a model-free setup. However, a model-free approach does not allow the system to be differentiable, which requires the use of high-variance gradient estimations. In this paper we introduce the Model based Adversarial Imitation Learning (MAIL) algorithm. A model-based approach for the problem of adversarial imitation learning.
We show how to use a forward model to make the system fully differentiable, which enables us to train policies using the (stochastic) gradient of $D$. Moreover, our approach requires relatively few environment interactions, and fewer hyper-parameters to tune. We test our method on the MuJoCo physics simulator and report initial results that surpass the current state-of-the-art.
\end{abstract}

\section{Introduction}
Learning a policy from scratch is often difficult. However, in many problems, there exists an expert policy which achieves satisfactory performance. We're interested in the scenario of imitating an expert. Imitation is needed for several reasons: \emph{Automation} (in case the expert is human), \emph{distillation} (e.g. if the expert is too expensive to run in real-time, \citep{rusu2015policy}), and \emph{initialization} (using an expert policy as an initial solution). In our setting, we assume sample trajectories $\{s_0,a_0,s_1,...\}_{i=0}^N$ of an expert policy $\pi_E$ are given, and the goal to train a new policy $\pi$ which imitates $\pi_E$ from demonstrations without access to the original reward signal $r_E$.

There are two main approaches for solving imitation problems.
The first, known as Behavioral Cloning (BC), directly learns the conditional distribution of actions over states $p(a|s)$ in a Supervised Learning (SL) fashion \citep{pomerleau1991efficient}. By providing constant supervision (i.e. a "dense" reward signal), BC overcomes fundamental difficulties of RL such as the credit assignment problem \citep{sutton1984temporal}. However, BC has its downsides as well. In oppose to Temporal Difference (TD) methods \citep{sutton1988learning} that incorporate information over time, BC methods are trained using single time-step state-action pairs $\{s_t,a_t\}$. This indifference to the dynamics of the environment makes BC methods susceptible to suffer compounding errors known as covariate shifts \citep{ross2010efficient, ross2011reduction}. On top of that, the sample complexity of BC methods is usually high, requiring a large amount of expert data that could be expensive to produce.

\par
The second approach is built out of two phases. In the first phase, called Inverse Reinforcement Learning (IRL), \citep{ng2000algorithms}, one tries to recover a reward signal under which the expert is uniquely optimal:
\begin{equation}
\mathbb{E} \big[ \sum_t \gamma^t \hat{r}(s_t,a_t) | \pi_E \big] \geq \mathbb{E} \big[ \sum_t \gamma^t \hat{r}(s_t,a_t) | \pi \big] \quad \forall \pi.
\end{equation}
After a reward signal $\hat{r}$ is obtained, the second phase is to apply standard RL techniques in order to maximize the discounted cumulative expected return $\mathbb{E}_\pi R = \mathbb{E}_\pi \big[ \sum_{t=0}^T{\gamma^t \hat{r}_t} \big]$.
However, IRL problems are usually hard to solve since the problem of recovering a reward signal from demonstrations is severely ill-posed \citep{ziebart2008maximum}. Instead of performing IRL, one can try other methods for building $\hat{r}$, but overall, synthesizing a reward signal like the one that was used by the expert is very challenging and requires extensive domain knowledge \citep{dorigo1998robot}.

\par
Generative Adversarial Networks (GAN) \citep{goodfellow2014generative} is a new approach for training generative models. With GAN, \textit{dense} supervision is provided in the form of a discriminator Neural Network (NN) $D$ that is trained to discriminate between the generative model $G$ and the expert's data. The unique form of guidance offered by GAN makes it appealing for other purposes besides its original intent; Image captioning \citep{mirza2014conditional}, and Video prediction \citep{mathieu2015deep} are some examples. 
More recently, a work named Generative Adversarial Imitation Learning (GAIL) \citep{ho2016generative}, has successfully applied the ideas of GAN for imitation learning in a model-free setup. They showed that this type of learning could alleviate problems like sample complexity or compounding errors, traditionally coupled with imitation learning.
\par
However, a model-free approach has its limitations. One of them is that the generative model can no longer be trained by simply backpropagating the gradients from the loss function defined over $D$. Instead, the model-free approach resorts to high-variance gradient estimations. In this work, we present a model-based version of adversarial imitation learning. We show that by using a forward model, the system can be easily trained end-to-end using regular backpropagation. More explicitly, the policy gradient can be derived directly from the gradient of the discriminator- the original core idea behind this type of learning. The resulting algorithm we propose processes entire trajectories with the objective of minimizing the total sum of discriminator probabilities along a path. In this way, we can train policies that are more robust and require fewer interactions with the environment while training.

\section{Background}
\subsection{Markov Decision Process}
Consider an infinite-horizon discounted Markov decision process (MDP), defined by the tuple $(S, A, P, r, \rho_0, \gamma)$, where $S$ is a finite set of states, $A$ is a finite set of actions, $P : S \times A \times S \rightarrow R$ is the transition probability distribution, $r : (S \times A) \rightarrow R$ is the reward function, $\rho_0 : S \rightarrow R$ is the distribution of the initial state $s_0$, and $\gamma \in (0, 1)$ is the discount factor. Let $\pi$ denote a stochastic policy $\pi : S \times A \rightarrow [0, 1]$, $R(\pi)$ denote its expected discounted reward: $\mathbb{E}_\pi R = \mathbb{E}_\pi \big[ \sum_{t=0}^T{\gamma^t \hat{r}_t} \big]$, and $\tau$ denote a trajectory of states and actions $\tau=\{s_0,a_0,s_1,a_1,...\}$.

\subsection{Imitation Learning}
Learning control policies directly from expert demonstrations, has been proven very useful in practice, and has led to satisfying performance in a wide range of applications. A common approach to imitation learning is to train a policy $\pi$ to minimize some loss function $l(s,\pi(s))$, under the discounted state distribution encountered by the expert: $d_\pi(s)=(1-\gamma)\sum_{t=0}^\infty \gamma^t p(s_t)$. This is possible using any standard SL algorithm:
$$\pi=\argmin_{\pi \in \Pi}  \, \mathbb{E}_{s \sim d_\pi} [l(s,\pi(s))],$$
where $\Pi$ denotes the class of all possible policies.
However, the policy's prediction affects the future state distribution, which violates the i.i.d assumption made by most SL algorithms. A slight deviation in the learner's behavior may lead it to a different state distribution than the one encountered by the expert, resulting in compounding errors.
\par
To overcome this issue, \cite{ross2010efficient} introduced the Forward Training (FT) algorithm that trains a non-stationary policy iteratively over time (one policy $\pi_t$ for each time-step). At time $t$, $\pi_t$ is trained to mimic $\pi_E$ on the state distribution induced by the previously trained policies $\pi_0,\pi_1, \pi_{t-1}$. This way, $\pi_t$ is trained on the actual state distribution it will encounter at inference.
However, the FT algorithm is impractical when the time horizon $T$ is large (or undefined), since it needs to train a policy at each time-step, and cannot be stopped before completion. The Stochastic Mixing Iterative Learning (SMILe) algorithm, proposed by the same authors, solves this problem by training a stochastic stationary policy over several iterations. SMILe starts with an initial policy $\pi_0$ that blindly follows the expert's action choice. At iteration $t$, a policy $\hat{\pi_t}$ is trained to mimic the expert under the trajectory distribution induced by $\pi_{t-1}$, and then updates $\pi_t=\pi_{t-1} + \alpha (1- \alpha)^{t-1} (\hat{\pi_t} - \pi_0)$. Overall, both the FT algorithm and SMILe gradually modify the policy from following the expert's policy to the learned one.

\subsection{Generative Adversarial Learning}
GAN suggests learning a generative model in an adversarial process, by phrasing it as a minimax two-player game with the following value function:
\begin{equation}
\label{eq:GAN}
\argmin_{G} \, {\argmax_ {D \in (0,1)} \, {\mathbb{E}_{x \backsim p_E}\ [\log D(x)] + \mathbb{E}_{z \backsim p_z} \big[ \log \big( 1-D(G(z)) \big) \big]}}.
\end{equation}
In this game, player $D$ is a differentiable function represented by a Neural Network (NN), with the objective of maximizing Eq.~\ref{eq:GAN}. $D$ maximizes the objective by learning to discriminate between the expert data, and data that the opponent player, $G$ (also modeled by an NN), generates on the fly.
$G$ in its turn, tries to minimize Eq.~\ref{eq:GAN}. He uses $D$ to define a loss function $l(z,\theta_g)=\log \big( 1-D_{\theta_g}(G(z))$, that when minimized increases the probability of $D$ to misclassify the data that $G$ generates. Eventually, $G$ learns to approximate the data distribution of $p_E(x)$ which is the desired goal.
\par
The main advantage of GAN on previous methods is that there is no need to train cumbersome models like RBM and DBN \citep{lee2009convolutional}. Instead, one can rely on standard backpropagation to train the system end-to-end. The discriminator trains by ascending its gradient of Eq.~\ref{eq:GAN}:
$$\nabla_{\theta_d} \frac{1}{m} \sum_{i=1}^m \Big[ \log D \big( x^{(i)} \big) + \log \Big( 1-D \big( G(z^{(i)} \big) \Big) \Big],$$
alternately while updating the generator by descending its gradient:
$$\nabla_{\theta_g} \frac{1}{m} \sum_{i=1}^m \log \Big( 1- D \big( G(z^{(i)}) \big) \Big).$$

Following GAN's success, GAIL suggested using the same idea for learning how to imitate an expert policy in a model-free setup. GAIL draws a similar objective function like GAN, except that now $p_E$ stands for the expert's joint distribution over state-action tuples:
\begin{equation}
\label{eq:GAIL}
\argmin_{\pi} \, {\argmax_ {D \in (0,1)} \, {\mathbb{E}_{\pi}[\log D(s,a)] + \mathbb{E}_{\pi_E} [\log(1-D(s,a))]} -\lambda H(\pi)},
\end{equation}
where $H(\lambda)\triangleq \mathbb{E}_{\pi}[-\log \pi (a|s)] $ is the causal entropy.
\par
While the optimization of the discriminator can still be done using backpropagation, this is not the case for the optimization of the generator (policy). Eq.~\ref{eq:GAIL} depends on $\pi$ indirectly through the first term: $\mathbb{E}_{\pi}[\log D(s,a)]$. The dependence is indirect since $\pi$ affects the data distribution, but do not appear in the objective itself. Assume that $\pi=\pi_\theta$, it's unclear how to differentiate Eq.~\ref{eq:GAIL} with respect to $\theta$. A common solution is to use the likelihood-ratio estimator, of which the popular REINFORCE algorithm \citep{williams1992simple}, is a special case:
\begin{equation}
\nabla_\theta \mathbb{E}_{\pi}[\log D(s,a)] \cong \mathbb{\hat{E}}_{\tau_i}[\nabla_\theta \log \pi_\theta(a|s) Q(s,a)],
\end{equation}
where $Q(\hat{s},\hat{a})$ is the score function of the gradient:
\begin{equation}
\label{eq:gail_obj}
Q(\hat{s},\hat{a})=\hat{\mathbb{E}}_{\tau_i}[\log D(s,a) \, | \, s_0=\hat{s},a_0=\hat{a}].
\end{equation}
\par
Although unbiased, the REINFORCE gradient estimation tends to have high variance, making it hard to work with even after applying variance reduction techniques \citep{ranganath2014black, mnih2014neural}. We claim that the reason is that the REINFORCE gradient discards the Jacobian matrix of the graph part downstream the stochastic unit. In the following we show how by including a forward model of the environment, the system can be differentiable end-to-end, allowing us to use the partial derivatives of $D: [\nabla_aD, \nabla_s D]$ when differentiating Eq.~\ref{eq:GAIL} with respect to $\theta$.
\par
Moreover, the reliance on the likelihood-ratio estimator makes the model-free approach demanding in the number of environment interactions.
Due to its high variability, the REINFORCE gradient requires running multiple trajectories at each time-step in order to get a good estimation of the score function $Q(\hat{s},\hat{a})$. In this paper we show how the model-based approach can reduce this demand. Avoiding using REINFORCE will also alleviate other technical issues such as resuming the environment at previously visited states, a troublesome request for some applications. 

\begin{figure}
\begin{center}
\centerline{\includegraphics[trim=0cm 1.2cm 0cm 0cm, clip, width=0.4\columnwidth]{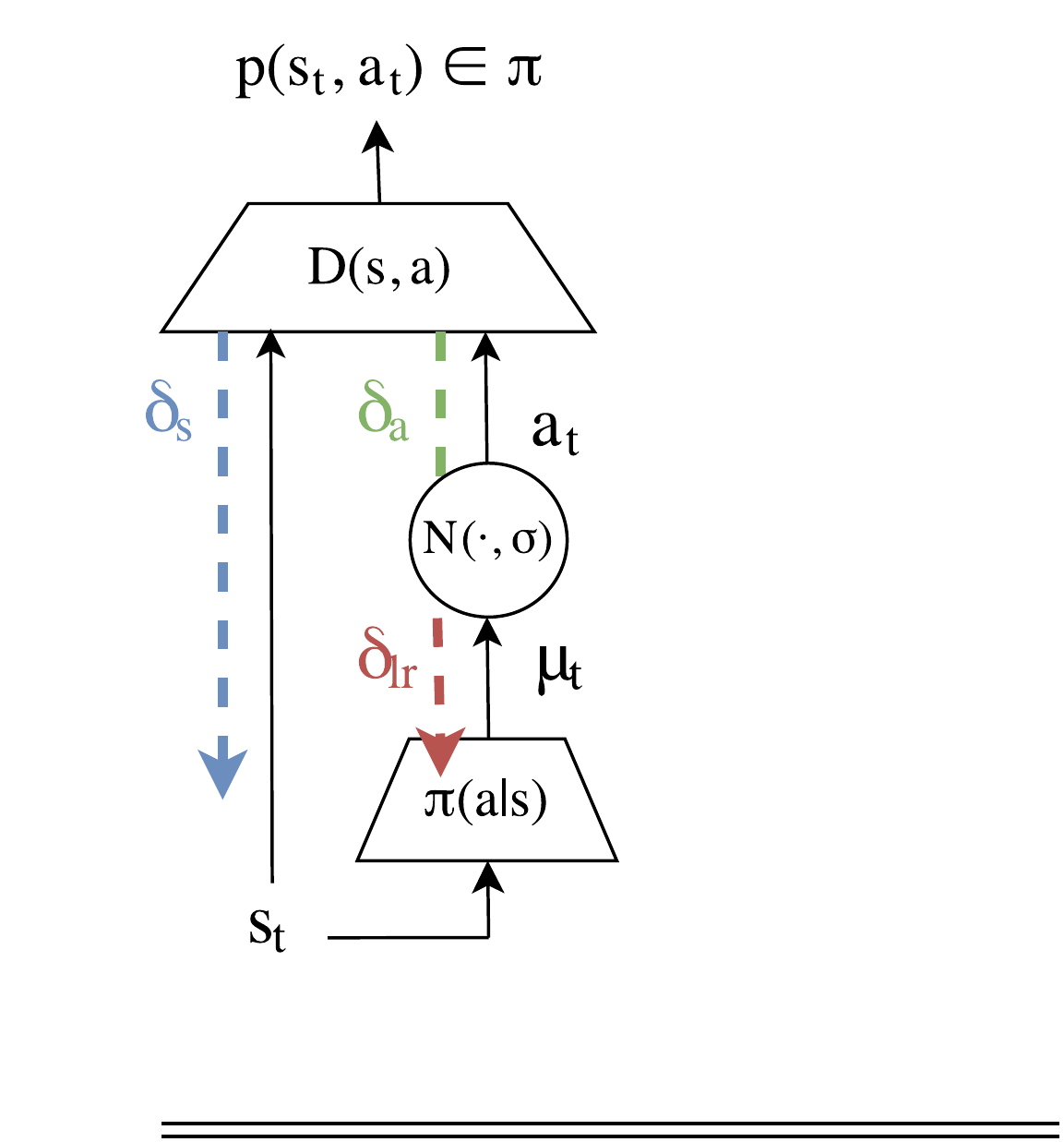}}
\caption{\textbf{Model-free adversarial imitation learning block diagram.} Given a state $s$, the policy outputs $\mu$ which is fed as the mean to a Gaussian sampling unit. An action $a$ is then sampled, and together with $s$ are fed into the discriminator network. In the backward phase the error message $\delta_a$, is \emph{blocked} at the stochastic sampling unit, therefore high-variance gradient estimation is used ($\delta_{lr}$). Meanwhile, the error message from the state input, $\delta_s$ is \emph{flushed} without being used at all.}
\label{gail_block_diagram}
\end{center}
\end{figure}

\section{Algorithm}
The discriminator network is trained to predict the conditional distribution: $D(s,a)=p(y|s,a)$ where $y=\{\pi_E,\pi\}$. I.e., $D(s,a)$ represents the probability that $\{s,a\}$ are generated by $\pi$ rather than $\pi_E$. Using Bayes rule and the law of total probability we get that:
$$D(s,a)=p(\pi|s,a)=\frac{p(s,a|\pi)p(\pi)}{p(s,a)}=\frac{p(s,a|\pi)p(\pi)}{p(s,a|\pi)p(\pi)+p(s,a|\pi_E)p(\pi_E)}=\frac{p(s,a|\pi)}{p(s,a|\pi)+p(s,a|\pi_E)} \, .$$
The last move is correct since the discriminator is trained on an even distribution of expert/generator examples, therefore we have that: $p(\pi)=p(\pi_E)=\frac{1}{2}$. 
Re-arranging and factoring the joint distribution we can write that:
$$D(s,a)=\frac{1}{\frac{p(s,a|\pi)+p(s,a|\pi_E)}{p(s,a|\pi)}}=\frac{1}{1+\frac{p(s,a|\pi_E)}{p(s,a|\pi)}}=\frac{1}{1+\frac{p(a|s,\pi_E)}{p(a|s,\pi)} \cdot \frac{p(s|\pi_E)}{p(s|\pi)}} \, \, .$$
Denoting $\varphi(s,a)=\frac{p(a|s,\pi_E)}{p(a|s,\pi)}$, and $\psi(s)=\frac{p(s|\pi_E)}{p(s|\pi)}$ we finally get that:
\begin{equation}
D(s,a)=\frac{1}{1+\varphi(s,a) \cdot \psi(s)} \, \, .
\end{equation}

Inspecting the derived expression we see that $\varphi(s,a)$ represents a \emph{policy likelihood ratio}, and $\psi(s)$ represents a \emph{state distribution likelihood ratio}. This interpretation suggests that the discriminator builds it logic by answering two questions. The first question relates to the \emph{state distribution}: "How likely is state $s$ under the distribution induced by $\pi_E$ vs. the one induces by $\pi$?", and the second question relates to the \emph{behavior}: "How likely is action $a$ given state $s$, under $\pi_E$ vs. $\pi$?".

We conclude that efficiently training a policy requires the learner to be aware of how his choice of actions affects the future \emph{state distribution} as well as how it affects the immediate \emph{behavior}.
In fact, inspecting the partial derivatives of $D$ with respect to $a$, and $s$, we see that the discriminator provides us with valuable information regarding the change in distribution, ${\partial \psi}/{\partial s}$, which is needed to make $\pi$ better like $\pi_E$:

\begin{equation}
\nabla_a D = - \frac{\varphi_a(s,a)\psi(s)}{(1+\varphi(s,a)\psi(s))^2} \quad , \quad \nabla_s D = - \frac{\varphi_s(s,a)\psi(s) + \varphi(s,a)\psi_s(s)}{(1+\varphi(s,a)\psi(s))^2} \, .
\end{equation}

Where $f_x$ stands for the partial differentiation $\partial f(x,y) / \partial x$.
A model-free solution is limited in its ability to use the partial derivatives of $D$ (see Figure~\ref{gail_block_diagram}). Next we show how full usage of $\nabla D = [\nabla_s D , \nabla_s D]$ is possible in the model-based approach.

\subsection{Re-parametrization of distributions (for using $\nabla_a D$)}
The first novelty we introduce is to re-write the stochastic policy using the re-parametrization trick, which permits us to compute derivatives of stochastic models. Assume that the policy is given by $\pi_\theta (a|s)=\mathcal{N}(a|\mu_\theta (s),\sigma^2_\theta(s))$, where $\mu,\sigma$ are deterministic functions. We can re-write it aModel-based Adversarial Imitation Learnings $\pi_\theta(a|s)=\mu_\theta(s) + \xi \sigma_\theta(s)$, where $\xi \sim \mathcal{N}(0,1)$. In this way we are able to get a Monte-Carlo estimator of the derivative of the expected discriminator probability of $(s,a)$ with respect to $\theta$:
\begin{equation}
\nabla_\theta \mathbb{E}_{\pi(a|s)}D(s,a)= \mathbb{E}_{\rho(\xi)} \nabla_a D(a,s) \nabla_\theta \pi_\theta (a|s) \cong \frac{1}{M}\sum_{i=1}^M{\nabla_a D(s,a)\nabla_\theta\pi_\theta (a|s)} \Big|_{\xi=\xi_i} \, .
\end{equation}

\subsection{Forward model (for using $\nabla_s D$)}
Using the partial derivative $\nabla_s D$ is a bit more tricky, and looking at the block diagram of the model-free approach (Figure~\ref{gail_block_diagram}), we understand why. The model-free approach treats the state $s$ as fixed and only tries to optimize the \emph{behavior}. Therefore, instead of viewing it as fixed, we suggest expressing $s$ as a function of the policy by setting: $s'=f(s,a)$, where $f$ is the forward model. This way, using the law of total derivative we get that:

\begin{equation}
\begin{split}
\nabla_\theta D(s_t,a_t) \Bigg|_{s=s_t,a=a_t} =&
\frac{\partial D}{\partial a} \frac{\partial a}{\partial \theta} \Bigg|_{a=a_t} + \frac{\partial D}{\partial s} \frac{\partial s}{\partial \theta} \Bigg|_{s=s_t} =
\\&
\frac{\partial D}{\partial a} \frac{\partial a}{\partial \theta} \Bigg|_{a=a_t} + \frac{\partial D}{\partial s} \Bigg(
\frac{\partial f}{\partial s} \frac{\partial s}{\partial \theta} \Bigg|_{s=s_{t-1}}+
\frac{\partial f}{\partial a} \frac{\partial a}{\partial \theta} \Bigg|_{a=a_{t-1}}
\Bigg)_.
\end{split} 
\end{equation}

\begin{figure}[t]
\begin{center}
\centerline{\includegraphics[trim=0cm 15.5cm 0cm 0cm, clip, width=\columnwidth]{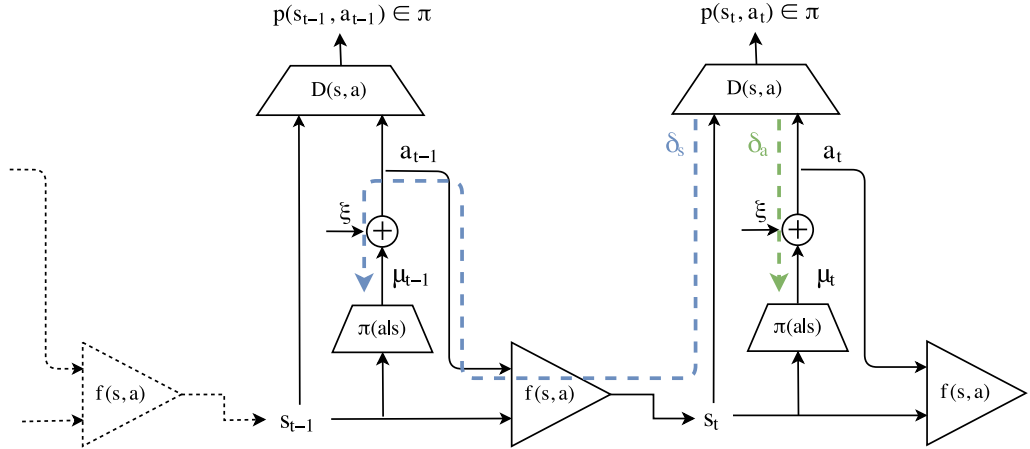}}
\caption{\textbf{Model-based adversarial imitation learning block diagram.} The green dashed arrow represents $\nabla_a D(s,a)$, and the blue one represents $\nabla_s D(s,a)$. Using the re-parametrization trick enables $\delta_a$ to backpropagate upstream the stochastic unit, which allow us to use the \emph{original error message} when calculating the gradient. In addition, by including a forward model we see how the error message $\delta_s$ backpropagates in time when calculating the gradient in the previous time step.}
\label{mgail_block_diagram}
\end{center}
\end{figure}

Since we have that $a=a_\theta$, we see that by considering a multi-step transition process, the error message of future \emph{state distributions} is accounted by earlier policy decisions. Figure~\ref{mgail_block_diagram} summarizes this idea.

\subsection{MAIL Algorithm}
We showed that effective imitation learning requires \emph{a)} to use a model, and \emph{b)} to process multi-step transitions instead of individual state-action pairs. This setup was previously suggested by \cite{shalev2016long} and \cite{heess2015learning}, who tried to maximize $R(\pi)$ by expressing it as a multi-step differentiable graph. Our method can be viewed as a variant of their idea when setting: $r(s,a)=-D(s,a)$. This way, instead of maximizing the total reward, we minimize the total discriminator beliefs along a trajectory.

\begin{minipage}{\textwidth}
Define $J(\theta)=\mathbb{E} \Big[ \sum_{t=0}\gamma ^t D(s_t,a_t) \big| \theta \Big]$ as the discounted sum of discriminator probabilities along a trajectory. Following the results of \cite{heess2015learning}, we write the derivatives of $J$ over a $(s,a,s')$ transition in a recursive manner:

\begin{equation}
\label{eq:d_objective_ds}
J_s=\mathbb{E}_{p(a|s)} \mathbb{E}_{p(s'|s,a)} \mathbb{E}_{p(\xi|s,a,s')}
\Bigg[ D_s + D_a \pi_s + \gamma J'_{s'}(f_s+f_a\pi_s) \Bigg] ,
\end{equation}
\begin{equation}
\label{eq:d_objective_dtheta}
J_\theta=\mathbb{E}_{p(a|s)} \mathbb{E}_{p(s'|s,a)} \mathbb{E}_{p(\xi|s,a,s')}
\Bigg[ D_a\pi_\theta + \gamma (J'_{s'}f_a\pi_\theta + J'_\theta)\Bigg] .
\end{equation}
The final gradient $J_\theta$ is calculated by applying Eq.~\ref{eq:d_objective_ds} and \ref{eq:d_objective_dtheta} recursively, starting from $t=T$ all the way down to $t=0$. The full algorithm is presented in Algorithm~\ref{mail_alg}.

\end{minipage}

\begin{algorithm}[]
\caption{\textbf{Model-based Adversarial Imitation Learning}}
\begin{algorithmic}[1]
\label{mail_alg}
  \STATE Given empty experience buffer $\mathcal{B}$
  \STATE \textbf{for} $trajectory=0$ \textbf{to} $\infty$ \textbf{do}
  \STATE $\quad$ \textbf{for} $t=0$ \textbf{to} $T$ \textbf{do}
  \STATE $\quad \quad$ Act on environment: $a=\pi(s,\xi; \theta)$
  \STATE $\quad \quad$ Push $(s,a,s')$ into $\mathcal{B}$
  \STATE $\quad$ \textbf{end for}
  \STATE $\quad$ train forward model $f$ using $\mathcal{B}$
  \STATE $\quad$ train discriminator model $D$ using $\mathcal{B}$
  \STATE $\quad$ set: $j'_s=0, j'_\theta=0$
  \STATE $\quad$ \textbf{for} $t=T$ \textbf{down to} $0$ \textbf{do}
  \STATE $\quad \quad$ $j_\theta=[D_a\pi_\theta + \gamma(j'_{s'}f_a\pi_\theta+ j'_\theta)] \big|_\xi$
  \STATE $\quad \quad$ $j_s=[D_s + D_a\pi_s + \gamma j'_{s'}(f_s+ f_a\pi_\theta)] \big|_\xi$
  \STATE $\quad$ \textbf{end for}
  \STATE $\quad$ Apply gradient update using $j^0_\theta$
  \STATE \textbf{end for}
\end{algorithmic}
\end{algorithm}

\section{Experiments}
We evaluate the proposed algorithm on two robotic challenges modeled by the MuJoCo physics simulator. Both tasks, \textbf{Hopper} and \textbf{Walker}, involve complex second order dynamics and direct torque control (further description provided below). We use the Trust Region Policy Optimization (TRPO) algorithm \citep{schulman2015trust} as the expert we wish to imitate. For each task, we produce four datasets of $\{25,18,11,4\}$ trajectories respectively, where each trajectory: $\tau=\{s_0, s_1,... s_N,a_N\}$ is of length $N=1000$.
\par
All networks comprise of 2 hidden layers with Relu non-linearity between, and are trained using the ADAM optimizer \citep{kingma2014adam}. Table~\ref{table:results} presents the total cumulative reward over a period of $N$ steps, measured using three different algorithms: BC, GAIL, and MAIL. The results for BC and GAIL are as reported in \citep{ho2016generative}. The MAIL algorithm achieves the highest reward for all dataset sizes while exhibiting performance comparable to the expert.
\subsection{Tasks:}
\subsubsection{Hopper}
The goal of the hopper task is to make a $2D$ planar hopper, with three joints and 4 body parts, hop forward as fast as possible. This problem has a 11 dimensional state space and a 3 dimensional action space that corresponds to torques at the joints. \\
\subsubsection{Walker}
The goal of the walker task is to make a $2D$ bipedal robot walk forward as fast as possible. The problem has a 17 dimensional state space and a 6 dimensional action space that corresponds to torque at the joints.

\begin{figure}
\centering
\subfigure[Hopper]{\label{fig:hopper}\includegraphics[trim={0.1cm 0 0cm 0},clip,height=50mm]{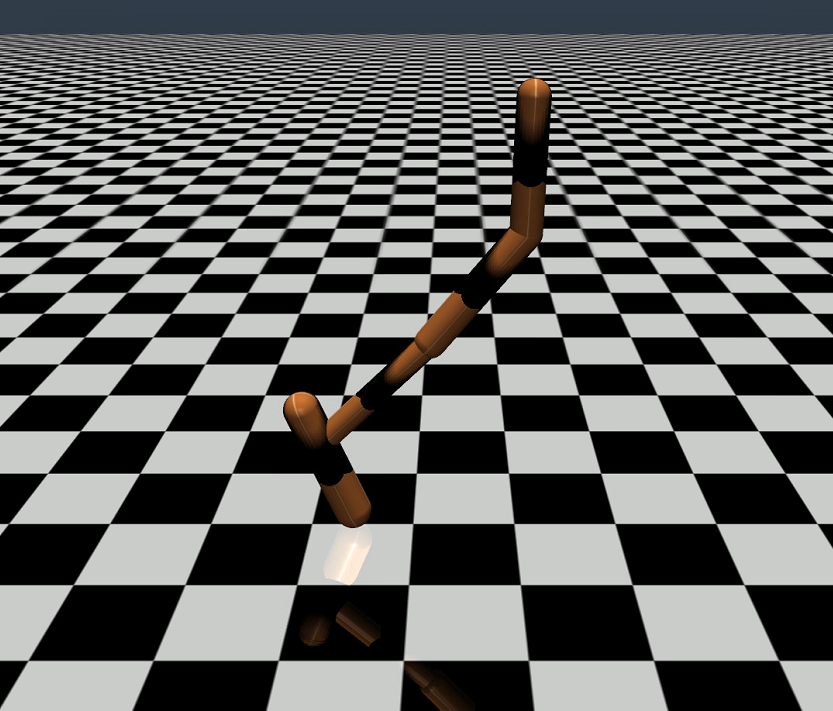}}
\subfigure[Walker]{\label{fig:walker}\includegraphics[trim={2cm 0 2cm 0},clip, height=50mm]{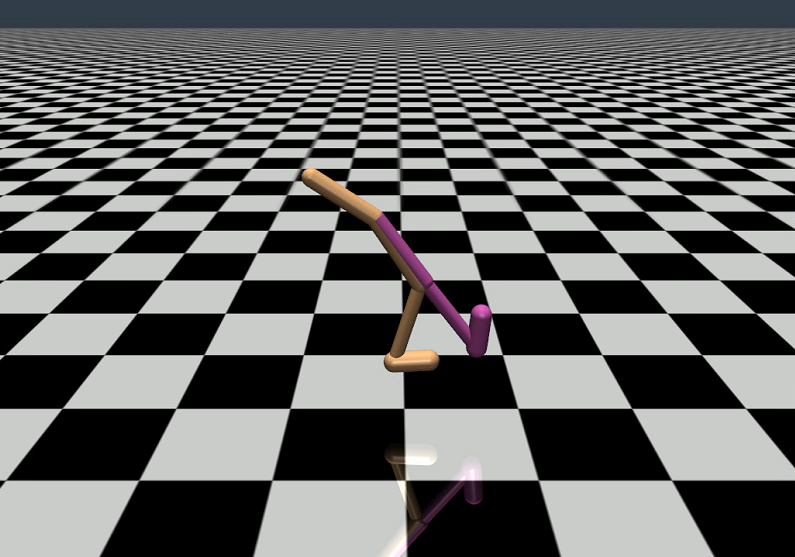}}
\caption{MuJoCo physics simulator}
\end{figure}

\begin{table}[]
\begin{center}
\begin{tabular}{@{}  lcccc @{}}    \toprule
Task & Dataset size & Behavioral cloning & GAIL & Ours  \\\midrule
 Hopper & 4 & $50.57 \pm 0.95     $  & $3614.22 \pm 7.17$ & $\bm{3669.53 \pm 6.09}$\\ 
        & 11 & $1025.84 \pm 266.86$  & $3615.00 \pm 4.32$ & $\bm{3649.98 \pm 12.36}$\\ 
        & 18 & $1949.09 \pm 500.61$  & $3600.70 \pm 4.24$ & $\bm{3661.78 \pm 11.52}$\\ 
        & 25 & $3383.96 \pm 657.61$  & $3560.85 \pm 3.09$ & $\bm{3673.41 \pm 7.73}$\\\midrule

 Walker & 4  & $32.18 \pm 1.25     $ & $4877.98 \pm 2848.37$ & $\bm{6916.34 \pm 115.20}$\\ 
        & 11 & $5946.81 \pm 1733.73$ & $6850.27 \pm 91.48$   & $\bm{7197.63 \pm 38.34}$\\ 
        & 18 & $1263.82 \pm 1347.74$ & $6964.68 \pm 46.30$   & $\bm{7128.87 \pm 141.98}$\\ 
        & 25 & $1599.36 \pm 1456.59$ & $6832.01 \pm 254.64$  & $\bm{7070.45 \pm 30.68}$\\        
 \hline
\end{tabular}
\caption{Learned policy performance}
\label{table:results}
\end{center}
\end{table}

\subsection{The Changing Distribution Problem}
Adversarial learning methods violate a fundamental assumption made by all SL algorithms, which requires the data to be i.i.d. The problem arises because the discriminator network trains on a changing data distribution produced by the training model. For the training to succeed, the discriminator must continually adapt to the changing distribution of the policy's data. In our approach, the problem is emphasized even more since not only the discriminator is affected but also the forward model. We train both $D(s,a)$ and $f(s,a)$ in a SL fashion using data that is constantly loaded into a replay buffer $\mathcal{B}$ \citep{lin1993reinforcement}.
\par
Because of the decreasing learning rate, the earliest seen examples have the strongest influence on the final solution. Assuming that the policy is initialized using some $\theta_0$ inducing data distribution different from the final one, it would be difficult to train $f$ because the system dynamics can be completely different at different areas of the state space (the same is true for $D$). A possible solution for this problem is to initialize the system with a BC training phase, where all three modules are trained directly from the expert data. A different solution is to \emph{restart} the learning multiple times along the training period by resetting the learning rate  \citep{loshchilov2016sgdr}. We tried both solutions without significant success. However, we believe that further research in this direction is needed.

\section{Discussion}
In this paper, we have presented a model based method for adversarial imitation learning. In comparison to the model-free approach, our method requires relatively few interactions with the environment, and fewer hyper-parameters to tune. However, our main advantage is that our approach enables us to use the partial derivatives of the discriminator when calculating the policy gradient. The downside of our approach is that it requires learning a forward model, which could be difficult in some problems. The accuracy of the forward model is crucial when backpropagating gradients recursively in time. An inaccurate model will lead to noisy gradients and will impede convergence.
\par
The system we propose comprises of multiple modules, which leads to many different training configurations. In our experiments, we tried several such configurations, which helped us to reach some conclusions.
We found that the discriminator network should be large ($\sim2x$) in comparison to the policy network it is guiding. Moreover, it should be trained with a large learning rate that slowly decades, because the discriminator needs to continually adapt to the changing distribution of the policy's data. We also found that the policy network should be trained more rapidly ($\sim3x$) than the discriminator or the forward model. We also found that adding noise to the expert data helps convergence, especially when working with few expert examples. Without noise the discriminator can always distinguish the expert from the policy, never being "satisfied" with the policy's distribution.
Finally, we note that the discriminator network holds valuable information that can be exploited for other purposes. The discriminator tells us in what parts of the state space the policy resembles the expert and where not. We can use this information for other goals besides its traditional use. For example, to prioritize training examples in the training phase, or as a confidence measure for the policy's performance at inference time.

\clearpage

\footnotesize
\bibliographystyle{plainnat}
\bibliography{paper_bib}
\end{document}